\documentclass{article}
\usepackage{spconf,amsmath,graphicx}

\usepackage{enumitem}
\usepackage{booktabs}
\setlist{nosep, leftmargin=14pt}

\usepackage[colorlinks,linkcolor=blue,anchorcolor=blue,citecolor=blue]{hyperref}
\usepackage{bbding}
\setlist{nosep, leftmargin=14pt}
\usepackage{hyperref}
\usepackage{mwe}

\usepackage{amssymb,amsfonts}
\usepackage{latexsym}
\usepackage{multirow}
\usepackage[table]{xcolor}
\usepackage{makecell}

\usepackage{mwe} % to get dummy images

\title{ASLseg: Adapting SAM in the Loop for Semi-supervised Liver Tumor Segmentation}
\name{Shiyun Chen$^{1\dagger}$, Li Lin$^{1,2\dagger}$, Pujin Cheng$^{1,2}$, Xiaoying Tang$^{1,3,*}$
\thanks{$^\dagger$ These authors contributed equally.}
\thanks{
$^*$Corresponding author: Dr. Xiaoying Tang (\url{tangxy@sustech.edu.cn})
}}
\address{$^1$Department of Electronic and Electrical Engineering, Southern University of Science and Technology,\\ Shenzhen, China\\
$^2$Department of Electrical and Electronic Engineering, The University of Hong Kong,\\Hong Kong SAR, China\\
$^3$Jiaxing Research Institute, Southern University of Science and Technology,\\Jiaxing, China}
\begin{document}
\maketitle
\begin{abstract}

Liver tumor segmentation is essential for computer-aided diagnosis, surgical planning, and prognosis evaluation. However, obtaining and maintaining a large-scale dataset with dense annotations is challenging. Semi-Supervised Learning (SSL) is a common technique to address these challenges. Recently, Segment Anything Model (SAM) has shown promising performance in some medical image segmentation tasks, but it performs poorly for liver tumor segmentation. In this paper, we propose a novel semi-supervised framework, named ASLseg, which can effectively adapt the SAM to the SSL setting and combine both domain-specific and general knowledge of liver tumors. Specifically, the segmentation model trained with a specific SSL paradigm provides the generated pseudo-labels as prompts to the fine-tuned SAM. An adaptation network is then used to refine the SAM-predictions and generate higher-quality pseudo-labels. Finally, the reliable pseudo-labels are selected to expand the labeled set for iterative training. Extensive experiments on the LiTS dataset demonstrate overwhelming performance of our ASLseg.

\end{abstract}

\section{Introduction}
\label{sec:intro}
\vspace{-0.2cm}
Liver cancer is one of the most common cancers, so accurately determining the location and size of tumor areas is crucial for preoperative treatment, intraoperative resection and postoperative recurrence prevention and control \cite{c26}. Liver tumor segmentation using deep learning technology can significantly improve the work efficiency and accuracy of surgeons. However, the mainstream image segmentation methods rely on a large amount of annotated data to train the model, while in the medical field, the annotation data of images are not only dependent on expert knowledge, but also limited by data cost and privacy protection \cite{c24,c25}. Compared with the natural image field, the public data sets available for training are very limited \cite{c1}, especially for liver tumors, which are diverse in type, with indistinct boundaries and strong anisotropy. With that being said, there are large amounts of unlabeled  abdominal CT images and how to fully exploit the power of those unlabeled images is of great importance \cite{c27}.

To address this issue, Semi-Supervised Learning (SSL) has been widely adopted in the medical image segmentation field \cite{cvpr1}, which can effectively leverage unlabeled data to enhance the generalization ability of the model. There are two common training strategies for SSL: (1) consistency regularization, which aims to enforce the consistency of the model outputs under different perturbations \cite{c2,c3}; (2) self-training, which aims to use the predictions of a low-cost model to generate and filter noisy pseudo-labels and use them with the labeled data for re-training \cite{c4}. However, due to the low quality of pseudo-labels, most semi-supervised paradigms are still far from being fully supervised counterpart. Therefore, how to generate more reliable pseudo-labels and narrow the performance gap between semi-supervised and fully supervised is a pressing challenge to be addressed.

In recent months, the prompt-based Segment Anything Model (SAM) \cite{c5} have attracted much attention for its powerful generalization ability. Existing studies have shown that fine-tuned medical SAM can be effectively applied to clinical medical image segmentation tasks \cite{c7,c8,c10}. However, the current medical SAM lacks the domain expertise of liver tumor segmentation, and it is difficult to achieve satisfactory performance by relying solely on its prediction as the segmentation result. In contrast, the SSL model trained on a specific dataset can better learn the key features of liver tumor, but its output prediction still has unreliable results. How to effectively exploit the strong generalization capabilities of SAM and adapt them to SSL settings remains to be explored.

In this paper, we propose a simple and effective framework, named ASLseg, which leverages the target edge-awareness capability of SAM and the domain-specific knowledge of a specialized model to generate higher-quality pseudo-labels. We also introduce an adaptation network to refine the low-quality pseudo-labels that are caused by the segmentation model’s inaccurate predictions. This framework addresses the aforementioned issues and achieves superior performance.  

Our contributions are summarized as follows: 
(1) We propose a novel semi-supervised framework, namely ASLseg. ASLseg adapts SAM to semi-supervised pipeline and can effectively combine the domain expertise of existing semi-supervised paradigms with the generic knowledge of SAM. (2) We develop a series of data augmentation strategies for simulating prediction inaccuracies in segmentation networks and use them to train an adaptation network that is effectively applied to refine pseudo-labels in the pipeline. (3) Extensive experiments are conducted on the LiTS dataset. ASLseg outperforms other compared methods, proving the validity and superiority of our proposed framework.
% \begin{itemize}
% \item 
% We propose a novel semi-supervised framework, namely ASLseg. ASLseg adapts SAM to semi-supervised pipeline and can effectively combine the domain expertise of existing semi-supervised paradigms with the generic knowledge of SAM.
% \item 
% We develop a series of data augmentation strategies for simulating prediction inaccuracies in segmentation networks and use them to train an adaptation network that is effectively applied to refine pseudo-labels in the pipeline.
% \item 
% Extensive experiments are conducted on the LiTS dataset. ASLseg outperforms other comparative methods, proving the validity and superiority of our proposed framework.

% \end{itemize}

\begin{figure}[tb]
\centering
\includegraphics[width=0.48\textwidth]{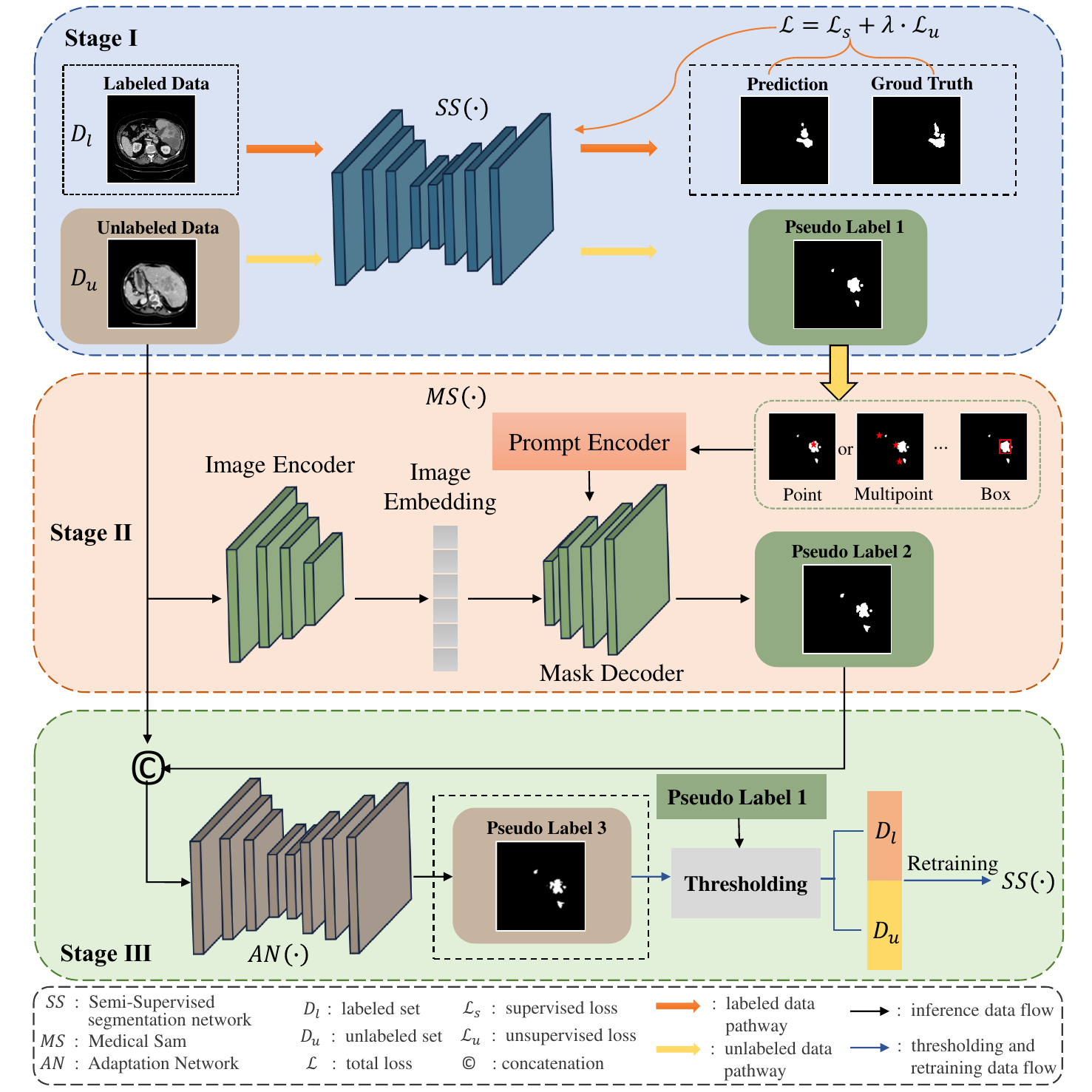}
\vspace{-0.6cm}
\caption{Overview of our proposed framework. Stage I is the SSL-specific training and inference module, Stage II is the fine-tuned medical SAM inference module, Stage III is the adaptation network inference module.} 
\label{fig:mainfig}
\vspace{-0.4cm}
\end{figure}

\vspace{-0.3cm}
\section{Methods}
\label{sec:format}
\vspace{-0.2cm}
As shown in Fig. \ref{fig:mainfig}, our ASLseg is a three-stage framework. In stage I, the segmentation model is trained in a specific semi-supervised manner using labeled data and pseudo-labels are generated by prediction on unlabeled data. In stage II, the labeled data is used to fine-tune the medicine SAM and combined with the pseudo-labels from stage one to make further predictions on the unlabeled data. In stage III, the pre-trained adaptation network using synthetic data refines the pseudo-labels generated by medical SAM and selects reliable pseudo-labels to rejoin the training of the semi-supervised segmentation model. This process will iterate multiple times until the best segmentation performance is achieved.

\vspace{-0.2cm}
\subsection{Overall Framework}
\vspace{-0.1cm}
\subsubsection{Semi-Supervised Learning}
\vspace{-0.1cm}
In our problem setting, given a training set $D_{ssl}$ = $\{D_l, D_u\}$ consisting of a labeled set $D_l$ = $\{(X_i, Y_i)_{i=1}^{N_t+N_{nt}}\}$ and a unlabeled set $D_u = \{(X_j)_{j=1}^{M_t+M_{nt}}\}$, where $X_i$/$X_j$ is the $i$/$j$ labeled/unlabeled image, $Y_i$ is the ground truth corresponding to the labeled image, $N_t$/$M_t$ is the number of labeled/unlabeled samples with tumors, $N_{nt}$/$M_{nt}$ is the number of labeled/unlabeled samples without tumors, the training objective is to train the segmentation model $\mathcal{SS}(\cdot)$. To utilize both the labeled and unlabeled data, the total loss $\mathcal{L}$ comprises the supervised loss $\mathcal{L}_{s}$ and the unsupervised loss $\mathcal{L}_{u}$:
\begin{equation}
\label{eqn:rdrop}
    \mathcal{L} = \mathcal{L}_{s} + \lambda \cdot \mathcal{L}_{u},
\end{equation}
where $\lambda$ is a regularization parameter to balance the supervised and unsupervised learning losses. Any suitable loss function for supervised semantic segmentation, such as cross entropy loss and dice loss, can be used as $\mathcal{L}_{s}$.

Taking R-Drop \cite{c11} as an example, for each training sample $X_i$, providing two forward passes of the network will result in two probability distributions with very small differences in the model predictions due to the presence of Dropout: $\mathcal{SS}_1(\cdot)$ and $\mathcal{SS}_2(\cdot)$. R-Drop utilizes the difference between these two prediction probabilities to constrain $SS_1$ and $SS_2$ by adding a symmetric Kullback-Leibler (KL) divergence loss of these two distributions: 
\begin{equation}
\begin{aligned}
\label{eqn:kl}
    \mathcal{L}_{KL} = \frac{1}{2} ( \mathcal{D}_{KL}(\mathcal{SS}_1(X_i) || \mathcal{SS}_2(X_i)) \\   
    + \mathcal{D}_{KL}(\mathcal{SS}_2(X_i) || \mathcal{SS}_1(X_i)) ).
\end{aligned}
\end{equation}
With cross entropy loss function of the two forward passes:
\begin{equation}
\label{eqn:nll}
    \mathcal{L}_{ce} = -\log \mathcal{SS}_1(X_i) - \log \mathcal{SS}_2(X_i),
\end{equation}
the final training loss function is: 
\begin{equation}
\label{eqn:ls}
    \mathcal{L}_{s} = \mathcal{L}_{ce} + \alpha \cdot \mathcal{L}_{KL},
\end{equation}
where $\alpha$ is a coefficient used to control $\mathcal{L}_{KL}$.

\vspace{-0.2cm}
\subsubsection{Medical SAM Fine-Tuning and Utilization}
\label{sam}
\vspace{-0.1cm}
Briefly (see the original article for a more refined formulation \cite{c10}), the medical SAM network $\mathcal{MS}(\cdot)$ uses labeled data with tumors as the training set $D_{sam} = \{(X_i, Y_i)_{i=1}^{N_t}\}$, fine-tuned using a combination of random and iterative click sampling strategies, which allows it to fully learn the tumor region feature knowledge. In the inference stage, we use random foreground points from the pseudo-labels predicted by the segmentation network as a single point prompt, to further segment the unlabeled data.

\begin{figure}[t]
\centering
\includegraphics[width=0.5\textwidth]{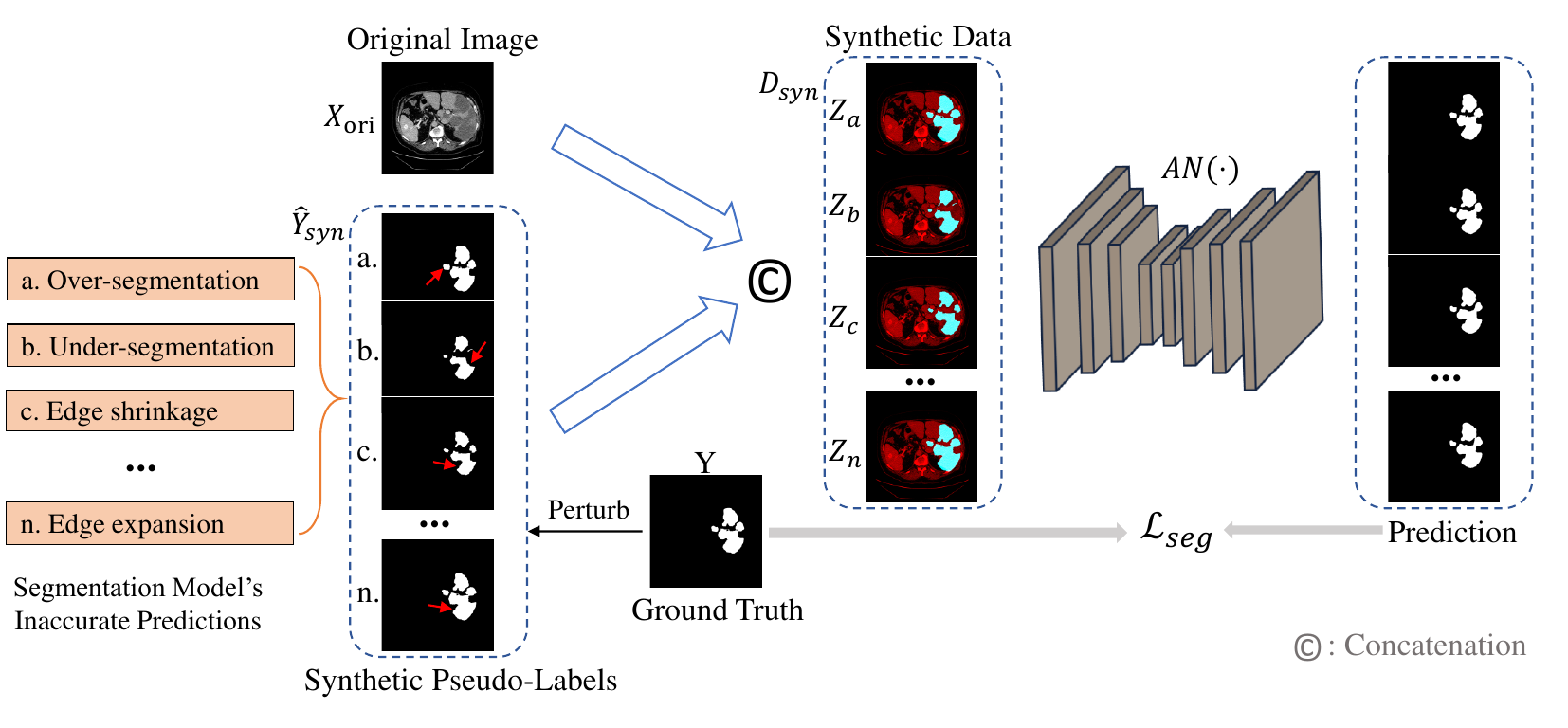}
\vspace{-0.7cm}
\caption{The training strategy of the adaptation network.} 
\label{fig:fig2}
\vspace{-0.3cm}
\end{figure}

\vspace{-0.2cm}
\subsubsection{Adaptation Network}
\label{2.1.3}
\vspace{-0.1cm}
 % We use the medical SAM model training set $D_{sam}$ to synthesize pseudo-labels to simulate various network segmentation inaccuracies. As shown in Fig. \ref{fig:fig2}, we perturb the ground truth masks with operations such as adding background noise, covering foreground regions, elastic transformation, expansion, erosion, etc., to generate pseudo-labels with different types of errors such as over-segmentation, under-segmentation, edge shrinkage, edge expansion, and so on. These operations are performed randomly with a probability threshold of 0.4. In addition, some all-black pseudo-labels are added to simulate the case where the model prediction is zero. 
 To enhance the performance of the adaptation network, we employed a training set augmentation method based on synthetic pseudo-labels. As shown in Fig. \ref{fig:fig2}, a series of operations were applied to increase the realism and diversity of the synthetic pseudo-labels, including adding background noise, covering foreground regions, elastic transformation, expansion, erosion, etc. on the ground truth masks $D_{sam}$. These operations were randomly executed, with a probability threshold of 0.4. Moreover, to address the potential miss-detection of the segmentation network, we also incorporated some all-black masks into the training set, simulating the scenario where the network prediction is zero.
 
 To enable the network to learn the key features of the target faster to compare and learn the differences between the pseudo-labels and the ground truth, the original images $X_{ori} = \{x_1,\cdots,x_R\}$ are concatenated with the corresponding synthetic pseudo-labels $\hat{Y}_{syn} = \{\hat{y}_1,\cdots,\hat{y}_R\}$ to construct a synthetic set $D_{syn} = \{(Z_k, Y_k)_{k=1}^{R}\}$ as the training set for the adaptation network  $\mathcal{AN}(\cdot)$, where $Z_k$ = $\operatorname{concat}(x_k,\hat{y}_k)$ is the synthetic data, $Y_k$ is the corresponding ground truth, and $R$ is the number of generated synthetic samples. This network structure adopts U-Net \cite{c14} architecture, and uses a linear combination of dice loss and cross entropy loss to supervise the prediction:
\begin{equation}
\label{rdrop}
    \mathcal{L}_{seg} = \mathcal{L}_{Dice}(Y,\mathcal{AN}(Z_k)) + \gamma \cdot \mathcal{L}_{ce}(Y,\mathcal{AN}(Z_k)),
\end{equation}
where $\gamma$ is a factor for balancing the Dice loss and cross-entropy loss contributions. In the inference stage, the SAM-prediction of unlabeled data is refined using the pre-trained adaption network.

\vspace{-0.2cm}
\subsection{Application of Liver Tumor Segmentation Task}
\label{task}
\vspace{-0.1cm}
% For the remaining unlabeled data $D_u$, we apply the segmentation model trained by R-Drop to these images to obtain segmentation proposals for each image.
It is assumed that the segmentation models $SS$ and $MS$ have been trained using the currently available labeled data. Moreover, it is assumed that for an unlabeled image $X_j \in D_u$, the segmentation region of the model $SS$ is used as a mask and a single point in the foreground region of the segmentation result is randomly selected as a positive prompt point. Then, these point coordinates are used as additional input information and fed into model $MS$ to obtain a set of segmentation predictions. Finally, this set of segmentation suggestions and the unlabeled image are concatenated and fed into model $AN$ to obtain the refined pseudo-annotations. 

The Dice Coefficient (DSC) index is used to evaluate the similarity of the predictions generated by segmentation model $SS$ and adaptation model $AN$, and set a threshold $\beta$ to determine whether their predictions are consistent. Then the pseudo-labels with high similarity and corresponding unlabeled images are selected and added to the labeled set, and fed together with the semi-supervised model for retraining, iterating until the optimal performance is achieved. Note that if different prompts from pseudo-labels are selected for input to the medical SAM (e.g., multipoint, box, or interaction prompts), different outputs can be produced, and thus our framework can be generalized to different input prompts. We will explore this further as future work.

\vspace{-0.3cm}
\section{EXPERIMENTS AND RESULTS}
\label{exp}
\vspace{-0.2cm}
\subsection{Dataset}
\label{data}
\vspace{-0.1cm}
We use the publicly available LiTS dataset \cite{c1}. We truncate the image intensity values of the CT scans to the range of [-82, 198] HU and normalize them, and then convert them into image slices (6548 with tumors and 1683 without tumors). For each slice ($512 \times 512$), the tumors with pixels less than 100 are excluded. We construct a semi-supervised dataset with the training and test sets ratio of 8:2, where 10\% of the data in the training set is used for validation during training. This dataset is used for fair comparison and validation of all methods and ablation experiments. In our dataset, the proportion of slices with tumors in both the training and test sets is 80\%.

For models trained with the SSL paradigm, the training set contains 6707 (10\% labeled and 90\% unlabeled) slices, the test set has 1524 slices. For medical SAM model fine-tuning, non-empty masks are required as prompt inputs during the training process, so 80\% of the slices with tumors are selected from the 10\% labeled data as the training set, and 80\% of the slices with tumors are selected from the semi-supervised test set as the test set. For adaptation network training, we use the synthetic data mentioned in sec. \ref{2.1.3} to train it.

\begin{figure*}[!t]
\centering
\includegraphics[width=1.0\textwidth]{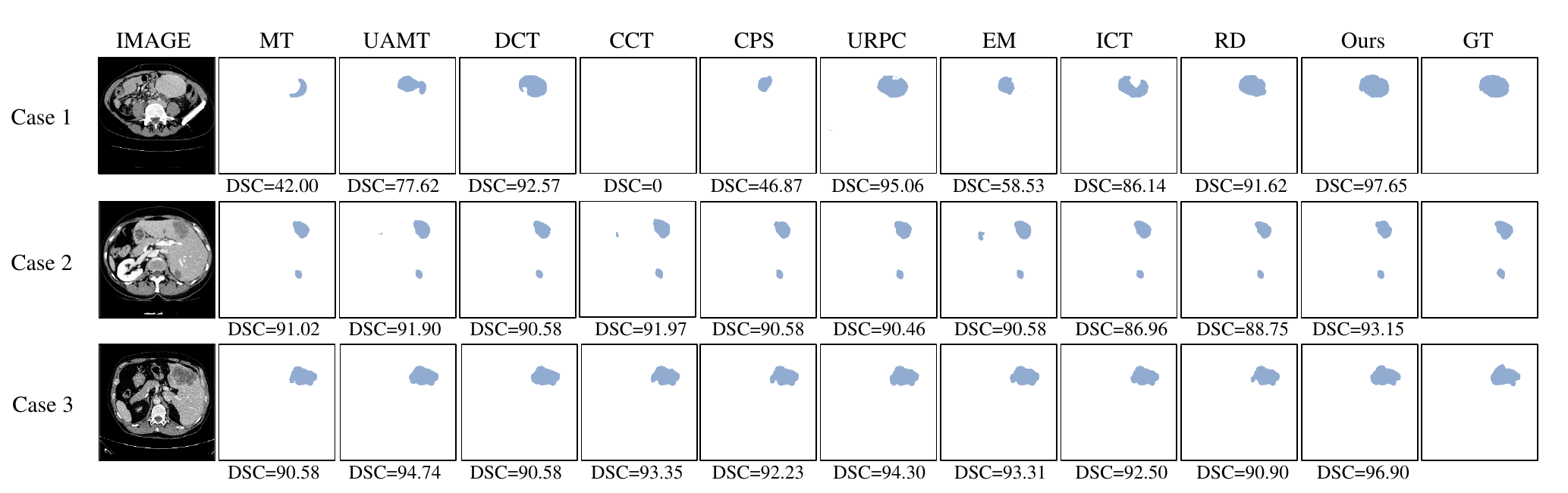}
\vspace{-0.7cm}
\caption{Visualize comparison results on the LiTS dataset.} 
\label{fig:fig3}
\vspace{-0.2cm}
\end{figure*}

\vspace{-0.2cm}
\subsection{Implementation Details}
\label{details}
\vspace{-0.1cm}
Since R-Drop achieves the best performance among SSL methods in the internal comparison experiments on the constructed dataset, we apply it to the proposed ASLseg framework to further demonstrate the effectiveness of the framework. For fairness sake, all methods chose U-Net as backbone, SGD as optimizer, 0.01 as the initial learning rate, and a polynomial scheduling strategy to update the learning rate. Batch size is set to 16. All methods performed 27000 iterations during training.

We chose Medical SAM Adapter \cite{c10} as the second stage model architecture and use U-Net as the basic architecture for adaption network. Batch size is set to 8. We perform 200 iterations during training. The threshold $\beta$ for determining the similarity between the predictions of two segmentation models for selecting reliable pseudo-labels is set to 0.9.

The entire pipeline is implemented by PyTorch using an NVIDIA GTX 2080Ti GPU.

\begin{table}[!t]
\vspace{-0.3cm}
  \centering
  {\caption{Comparison with existing SSL method on the LiTS dataset. The best results are highlighted in bold, and the second-best results are underlined. FS denotes fully-supervised learning (U-Net).}\label{table1}}

% \vspace{0.1cm}
  % \renewcommand{\arraystretch}{1.05}
  % \setlength{\tabcolsep}{1mm}
  \renewcommand\arraystretch{1.08}
  \setlength{\tabcolsep}{1.2mm}
  \renewcommand\theadgape{\Gape[1.8mm][0mm]}
  \resizebox{1.0\linewidth}{!}{
  \begin{tabular}{lccccc}
      \specialrule{0.1em}{0pt}{0pt}
\multirow{2}{*}{\makecell[l]{Method}}
& \multicolumn{5}{c}{LiTS} \\ 
\Xcline{2-6}{0.05em}
& DSC$\uparrow$           & JAC$\uparrow$          & SE$\uparrow$         & SP$\uparrow$    
& PRE$\uparrow$                   \\
\specialrule{0.1em}{0pt}{0pt}
MT (2017) \cite{c16}                 & 71.04$\pm$0.30             &  0.61$\pm$0.29           &  69.97$\pm$0.31           &   89.91$\pm$0.30           &\underline{76.74$\pm$0.31}                    \\
UAMT (2019)  \cite{c17}         &  71.67$\pm$0.30             &\underline{0.62$\pm$0.29}           &  71.33$\pm$0.31           &   90.77$\pm$0.28                           &76.06$\pm$0.31           \\
DCT (2019)  \cite{c18}                   & 69.63$\pm$0.31               & 0.60$\pm$0.30                                 &   68.54$\pm$0.31         & 69.93$\pm$0.32              &  73.75$\pm$0.32           \\
CCT (2020)  \cite{c19}                   &   71.11$\pm$0.29          &    0.61$\pm$0.28        &\underline{73.38$\pm$0.31}            &  90.91$\pm$0.28                   &  73.85$\pm$0.30            \\
CPS (2022)  \cite{c4}                  & 71.36$\pm$0.30              &  0.62$\pm$0.28           &   73.10$\pm$0.32                      &   90.55$\pm$0.29        &  74.17$\pm$0.30             \\
URPC (2021)  \cite{c2}                     &  69.12$\pm$0.31             &  0.60$\pm$0.30          &  71.09$\pm$0.33                       &   88.63$\pm$0.31         &  71.82$\pm$0.32    \\
EM (2019)  \cite{c22}               &   70.41$\pm$0.31            & 0.61$\pm$0.29           & 70.96$\pm$0.32            &  90.27$\pm$0.29                  &  74.02$\pm$0.32      \\
ICT (2022)  \cite{c23}                &  71.15$\pm$0.31             &  0.62$\pm$0.29           &   70.96$\pm$0.32          & 89.71$\pm$0.30                      &  75.51$\pm$0.31     \\
RD (2021)  \cite{c11}                     & \underline{72.16$\pm$0.29}          & 0.62$\pm$0.28            & 71.95$\pm$0.31          &\underline{90.99$\pm$0.28}           &\textbf{77.07$\pm$0.30} \\
Ours (ASLseg)      & \textbf{74.28$\pm$0.27}             & \textbf{0.65$\pm$0.27}                                    & \textbf{77.57$\pm$0.28}          & \textbf{94.73$\pm$0.22}                  & 75.50$\pm$0.29                    \\
FS\cellcolor{gray!30}                   & 76.16$\pm$0.28\cellcolor{gray!30}        & 0.67$\pm$0.27\cellcolor{gray!30}       & 74.56$\pm$0.29\cellcolor{gray!30}            &  91.37$\pm$0.27\cellcolor{gray!30}        &  81.27$\pm$0.28\cellcolor{gray!30}  \\        
        \specialrule{0.1em}{0pt}{0pt}

  \end{tabular}
  }
% \vspace{-0.3cm}
\end{table}

% \vspace{-0.5cm}
\begin{table}[!t]
\vspace{-0.3cm}
  \centering
      {\caption{Ablation study on the three key elements. SS denotes the segmentation model obtained by semi-supervised training, MS denotes the medical SAM, AN denotes the adaption network.}\label{table2}}
  % \vspace{0.1cm}
  % \renewcommand{\arraystretch}{1.05}
  % \setlength{\tabcolsep}{1mm}
  \renewcommand\arraystretch{1.08}
  \setlength{\tabcolsep}{1.2mm}
  \renewcommand\theadgape{\Gape[1.8mm][0mm]}
  \resizebox{1.0\linewidth}{!}{
  \begin{tabular}{ccc|ccccc}
\specialrule{0.1em}{0pt}{0pt}
 SS         & MS        & AN          & DSC$\uparrow$           & JAC$\uparrow$          & SE$\uparrow$         & SP$\uparrow$               & PRE$\uparrow$                   \\
\specialrule{0.1em}{0pt}{0pt}
 \checkmark              &            &              & 72.16$\pm$0.29          & 0.62$\pm$0.28            & 71.95$\pm$0.31          &90.99$\pm$0.28                 &   \textbf{77.07$\pm$0.30}                    \\
\checkmark             & \checkmark           &              &   72.83$\pm$0.28                  &   0.63$\pm$0.27            &   75.76$\pm$0.30         &93.68$\pm$0.24  &74.67$\pm$0.28  \\
\checkmark             &            &\checkmark              &   74.15$\pm$0.28                 &    \textbf{0.65$\pm$0.28}            &   76.67$\pm$0.28         &94.02$\pm$0.23  &76.00$\pm$0.29  \\
\checkmark             & \checkmark            & \checkmark                       &  \textbf{74.28$\pm$0.27}              & 0.65$\pm$0.27         &  \textbf{77.57$\pm$0.28}         &  \textbf{94.73$\pm$0.22}                & 75.50$\pm$0.29  \\ 
\specialrule{0.1em}{0pt}{0pt}
  \end{tabular}
  }
\vspace{-0.2cm}
\end{table}

\vspace{-0.2cm}
\subsection{Comparison and Ablation Experiments}
\label{cpmparison}
\vspace{-0.1cm}
To verify the effectiveness of our proposed strategy, we compare the results of our framework with fully supervised and nine SSL methods. For all competing methods, we use the official hyperparameter settings. 

Table \ref{table1} lists the quantitative results of different methods. When only 10\% of the data is labeled, our proposed framework outperforms nine SSL methods on four evaluation metrics, demonstrating the importance of mining information from unlabeled data. While fully supervised methods perform best in terms of accuracy, they require several times more training data than semi-supervised methods and only a fraction of the detected tumors are real tumors. However, misdetection and underdetection are usually more unwelcome in practical clinical medical diagnosis, as it may lead to inefficiency and malignant tumors being overlooked. Therefore, in addition to high accuracy, sensitivity and specificity need to be as high as possible. The results of our proposed method demonstrate that using SAM and the adaptation network can generate more reliable pseudo-labels and utilize unlabeled data more efficiently than existing methods. 

To evaluate the effectiveness of several key components in ASLseg, we conduct ablation studies in Table \ref{table2}. By adding two stages to refine pseudo-labels, the proposed method further improves DSC and achieves the best performance. In particular, our method significantly reduces the false positive and false negative rates while maintaining high DSC.

Representative visualization results are illustrated in Fig. \ref{fig:fig3}. As the advantages of the dedicated segmentation model and the general segmentation model are fully utilized, the boundaries of the tumor region can be identified more accurately, thus the significance of the components in the ASLseg is determined, and it is demonstrated that the ASLseg has a better pseudo-label refinement ability. 

\vspace{-0.3cm}
\section{CONCLUSION}
\label{sec:majhead}
\vspace{-0.2cm}
In this paper, we propose a novel semi-supervised liver tumor segmentation framework, which introduces SAM into SSL and also adds an adaption network, fully utilizing the unlabeled data. These newly introduced components can improve the quality and reliability of the generated pseudo-labels. Through extensive quantitative and qualitative experiments, our proposed ASLseg method achieves state-of-the-art performance on the widely used LiTS dataset, validating the effectiveness of our framework.

% \vspace{-0.2cm}
\section{Acknowledgments}
\label{sec:acknowledgments}
% \vspace{-0.2cm}
This study was supported by the Shenzhen Basic Research Program (JCYJ20190809120205578); the National Natural Science Foundation of China (62071210); the Shenzhen Science and Technology Program  (RCYX20210609103056042); the Shenzhen Basic Research Program (JCYJ2020092515384 7004); the Shenzhen Science and Technology Innovation Committee Program (KCXFZ2020122117340001).

% References should be produced using the bibtex program from suitable
% BiBTeX files (here: strings, refs, manuals). The IEEEbib.bst bibliography
% style file from IEEE produces unsorted bibliography list.
% ------------------------------------------------------------------------- 
% \tiny \scriptsize \footnotesize
% \small \normalsize \large Large
% \LARGE \huge \Huge \tiny 改变参考文献字体大小
% \vspace{-0.1cm}
\bibliographystyle{IEEEbib}
\bibliography{strings,refs}

\end{document}